\documentclass{article}

\PassOptionsToPackage{numbers,compress}{natbib}
\usepackage[preprint]{neurips_2026}
\makeatletter
\renewcommand{\@notice}{}
\makeatother

\usepackage[utf8]{inputenc}
\usepackage[T1]{fontenc}
\usepackage{hyperref}
\usepackage{url}
\usepackage{graphicx}
\usepackage{wrapfig}
\usepackage{array}
\usepackage{booktabs}
\usepackage{longtable}
\usepackage{multirow}
\usepackage{amsfonts}
\usepackage{nicefrac}
\usepackage{microtype}
\usepackage{xcolor}
\usepackage{xspace}
\usepackage{amsmath}
\usepackage{placeins}
\usepackage{cleveref}   

\crefname{figure}{Fig.}{Figs.}
\Crefname{figure}{Figure}{Figures}
\crefname{table}{Table}{Tables}
\Crefname{table}{Table}{Tables}
\crefname{section}{Sec.}{Secs.}
\Crefname{section}{Section}{Sections}
\crefname{appendix}{Appendix}{Appendices}
\Crefname{appendix}{Appendix}{Appendices}
\crefname{equation}{Eq.}{Eqs.}
\Crefname{equation}{Equation.}{Equations}

\newcommand{\latin}[1]{\textit{#1}\xspace} 
\newcommand{\eg}{\latin{e.g.}}

\newcommand{\vs}{\latin{vs.}}

\newcommand{\alg}{GeoViSTA\xspace}
\newcommand{\papertitle}{GeoViSTA: Geospatial Vision-Tabular Transformer for Multimodal Environment Representation}

\title{\textbf{\papertitle}}
\author{
{\normalfont Yuhao Liu\textsuperscript{1}}
\and
Sadeer Al-Kindi\textsuperscript{2}
\and
Ashok Veeraraghavan\textsuperscript{1}
\and
Guha Balakrishnan\textsuperscript{1}
\and
\textsuperscript{1}Department of Electrical and Computer Engineering, Rice University, Houston, TX 77005\\
\textsuperscript{2}Center for Cardiovascular Computational and Precision Health, Department of Cardiology,\\
DeBakey Heart and Vascular Center, Houston Methodist, Houston, TX 77030\\
{\tt\small \{yuhao.liu, guha, vashok\}@rice.edu, sal-kindi@houstonmethodist.org}
}
\date{}

\begin{document}

\maketitle

\begin{abstract}
Large-scale pretraining on Earth observation imagery has yielded powerful representations of the natural and built environment. However, most existing geospatial foundation models do not directly model the structured socioeconomic covariates typically stored in tabular form. This modality gap limits their ability to capture the complete total environment, which is critical for reasoning about complex environmental, social, and health-related outcomes. In this work, we propose \alg (Geospatial Vision-Tabular Transformer), a vision-tabular architecture that learns unified geospatial embeddings from co-registered gridded imagery and tabular data. \alg utilizes bilateral cross-attention to exchange spatial and semantic information across modalities, guided by a geography-aware attention mechanism that aligns continuous image patches with irregular census-tract tokens. We train \alg with a self-supervised joint masked-autoencoding objective, forcing it to recover missing image patches and tabular rows using local spatial context and cross-modal cues. Empirically, \alg's unified embeddings improve linear probing performance on high-impact downstream tasks, outperforming baselines in predicting disease-specific mortality and fire hazard frequency across held-out regions. These results demonstrate that jointly modeling the physical environment alongside structured socioeconomic context yields highly transferable representations for holistic geospatial inference.
\end{abstract}

\section{Introduction}
\label{sec:intro}
Geospatial environmental data are increasingly used to model high-impact applications such as social vulnerability, environmental risk, disaster response, and public health outcomes~\citep{ravuri2021skilful,liu2025downscalinga,rathje2017designsafe,aminidebris,rajagopalan2018air,al2020environmental,agarwal2025general}. The relevant signals for these tasks, however, are distributed across fundamentally different data structures. Natural and built environmental signals are commonly represented as continuous gridded imagery derived from remote sensing platforms~\citep{gorelick2017google,chen2022rainnet,veillette2020sevir}, whereas socioeconomic, demographic, and vulnerability-related variables are typically represented as structured tabular attributes aggregated over irregular administrative geographies such as counties or census tracts, often derived from survey-based and administrative data collection methods.

\begin{figure}[t]
\centering
\includegraphics[width=\linewidth]{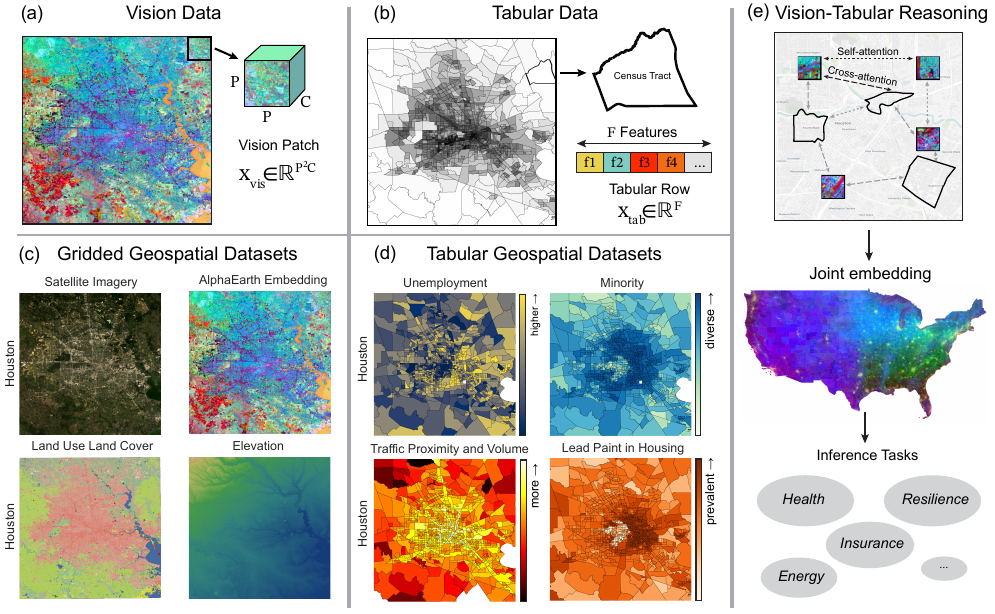}
\caption{\textbf{Vision--tabular geospatial data and reasoning}. (a-b) Vision and tabular datasets have fundamentally different structures: vision data are continuous, gridded feature images, while tabular data represent features over irregular geographical regions. (c-d) Examples of geospatial vision and tabular signals. Vision signals describe apparent facets of the natural and built environment, whereas tabular datasets provide human-related socioeconomic data. (e) This study proposes \alg, a transformer enabling visual-tabular reasoning, producing a joint embedding space intended to support a broad range of downstream geospatial inference tasks for multimodal environments.}
\label{fig:opening_fig}
\vspace{-0.6em}
\end{figure}

Multimodal geospatial reasoning therefore requires integrating fundamentally different data structures (\cref{fig:opening_fig}a-b). Vision-based geospatial data are organized as continuous gridded feature maps describing the natural and built environment, while tabular datasets encode social, economic, and institutional context through structured attributes linked to irregular geographic regions (\cref{fig:opening_fig}c-d). These complementary modalities capture distinct but interconnected determinants of environmental exposure, vulnerability, and downstream health outcomes.

As geospatial analysis shifts from raw data repositories to reusable embeddings, this modality gap becomes a representation-learning problem. Whereas early large-scale analyses~\citep{naik2017computer,anderson1976land} relied on substantial bespoke modeling and feature engineering directly on raw data repositories, recent geospatial foundation models~\cite{brown2025alphaearth,klemmer2024satclip} treat the Earth's surface as a continuous neural representation field, encoding location and time into compact embeddings that provide transferable semantics for downstream inference. However, this progress has largely centered on gridded Earth-observation data~\cite{brown2025alphaearth, klemmer2024satclip, szwarcman2025prithvieo20, herzog2025olmoearth, jakubik2025terramind, guo2024skysense, cong2022satmae}: these models capture natural and built environment signals, but do not directly model the structured tabular attributes that encode socioeconomics, vulnerability, and risk exposures.

The central challenge for reasoning over the multimodal geospatial environments is therefore not merely to learn better visual embeddings, but to align gridded neural fields with complementary tabular attributes defined over irregular geographical boundaries (\eg, cities, counties, and census tracts~\footnote{Census tracts are small statistical subdivisions used by the U.S. Census Bureau, roughly corresponding to neighborhood-scale geographic units with an average population of $4,000$.}). This introduces two technical requirements. First, cross-modal feature fusion must account for irregular geometries and spatial proximity between vision and tabular inputs. Second, the joint representation space must be learned in a scalable, task-agnostic, and self-supervised manner.

 To address these challenges, we propose \alg (\textbf{Geo}spatial \textbf{Vis}ion-\textbf{Ta}bular Transformer), a vision-tabular architecture for joint geospatial representation learning (\cref{fig:opening_fig}e). \alg utilizes bilateral cross-attention to exchange spatial and semantic information across modalities. Crucially, we implement a geography-aware attention mechanism that biases token interactions based on spatial distance, encouraging localized reasoning. We train \alg as a joint masked autoencoder (MAE)~\citep{he2021masked}. By randomly masking out vision patches and entire tabular rows, we force the model to reconstruct the missing elements using both spatial context and cross-modal cues. This formulation provides a simple, effective, and scalable self-supervised objective, and requires no labeled data.

We demonstrate \alg by learning joint embeddings over the Contiguous United States (CONUS), fusing gridded AlphaEarth~\cite{brown2025alphaearth} visual representations with the tabular Climate Vulnerability Index (CVI)~\citep{cviabout}. We experimentally show that jointly modeling the physical environment and structured socioeconomic context yields highly transferable geospatial representations. Specifically, \alg embeddings improve linear-probe prediction of high-impact downstream variables on random held-out counties, such as disease-specific mortality rates and fire hazards, outperforming unimodal and feature-concatenation baselines. We further test regional extrapolation skill by withholding Washington state from training, where \alg achieves the strongest mortality-rate linear-probe performance. Furthermore, qualitative analysis reveals that \alg's attention layers exhibit locally meaningful neighborhood focus, and its learned feature space captures semantically coherent spatial patterns across the continent. Ultimately, this work provides a principled and effective foundation for holistic geospatial reasoning.

\section{Related Work}
\label{sec:related-work}

\textbf{Masked pretraining.} Masked autoencoders (MAEs)~\citep{he2021masked} provide a scalable objective for learning visual representations by reconstructing missing input from context. Multimodal MAE variants~\citep{ebrahimi2023lanistr,du2024tip} extend this by encouraging models to recover missing information using cross-modal cues. We adapt the MAE principles to geospatial vision-tabular data by jointly masking image patches and tabular rows. Unlike generic multimodal masking, our objective is geographically grounded: the model must infer missing elements using nearby spatial and cross-modal evidence, forcing the learned representations to encode both the physical environment and structured socioeconomic context.

\textbf{Geospatial representation learning.} Recent geospatial foundation models have demonstrated that rasterized Earth observations (EO) yield broad geospatial semantics. SatMAE~\citep{cong2022satmae} adapts MAE to temporal and multispectral EO, while SatCLIP~\citep{klemmer2024satclip} aligns EO-derived features with geographic coordinates via contrastive learning. Subsequent models~\citep{brown2025alphaearth,herzog2025olmoearth,guo2024skysense} further scale this pretraining across sensors, time, and geographies. Complementary to EO-centric models, PDFM~\citep{agarwal2025general} learns population-dynamics embeddings from aggregated geo-indexed signals such as maps, busyness, search trends, weather, and air quality over postal-code/county graphs. However, these approaches either focus on gridded EO or aggregate heterogeneous signals into location-level feature vectors, rather than directly aligning continuous image patches with structured tabular attributes defined over irregular administrative regions. \alg addresses this gap by jointly modeling gridded imagery alongside irregular tabular geographies through geography-aware cross-attention.

\textbf{Vision-tabular transformers.} In the tabular domain, transformer architectures have established that column-wise tokenization and attention yield highly competitive representations~\citep{huang2020tabtransformer,gorishniy2021revisiting}. Recent multimodal extensions bridge vision and tabular data via contrastive alignment~\citep{hager2023best} or MAE~\citep{ebrahimi2023lanistr,du2024tip}. However, these models are strictly designed for sample-level prediction, rigidly pairing a single image to a single tabular record. They are not designed to handle irregular geospatial boundaries, neighborhood structures, or the complex many-to-many relationships between continuous image patches and diverse tabular census tracts. \alg overcomes this limitation by explicitly modeling spatial proximity and geometry across modalities.

\section{Background} 
\label{sec:background}

\textbf{Vision masked autoencoder.} Masked autoencoders (MAEs) learn representations by removing parts of an input and training an encoder-decoder to reconstruct the missing content from the visible context~\citep{he2021masked}. For vision data, MAEs are typically implemented using a Vision Transformer (ViT)~\citep{dosovitskiy2021image}. Given an image $I \in \mathbb{R}^{C \times H \times W}$ (with channels $C$, height $H$, and width $W$), we partition it into $N_v = (H/P)(W/P)$ non-overlapping $P \times P$ patches and flatten them into a sequence $S \in \mathbb{R}^{N_v \times P^2C}$. A linear projection $f_p:\mathbb{R}^{P^2C}\rightarrow\mathbb{R}^{D}$ maps each patch to a $D$-dimensional token, yielding the embedded sequence $S' \in \mathbb{R}^{N_v \times D}$. During training, we mask a large fraction of these tokens. The ViT encoder processes only the visible tokens (augmented with positional embeddings). The decoder then takes these encoded tokens, along with learned mask tokens at the dropped positions, to predict the raw pixel values of the full sequence. Finally, the predicted patches are unpatchified into a reconstructed image $\hat{I}$, and the model is optimized using the mean squared error (MSE) between $I$ and $\hat{I}$ over the masked patches only.

\textbf{Tabular transformer.} Transformers adapt to tabular data by treating individual feature values as tokens~\citep{gorishniy2021revisiting}. For a tabular sample $x \in \mathbb{R}^{F}$ with $F$ features (columns), a tokenizer maps each scalar value $x_j$ to a $D$-dimensional token via a feature-specific embedding function $f_j$ and bias $b_j$, such that $s'_j=f_j(x_j)+b_j$. Stacking these tokens produces a sequence $S' \in \mathbb{R}^{F \times D}$, directly analogous to the embedded patch sequence in a ViT. The tabular transformer then applies self-attention across this sequence to learn contextualized representations that capture inter-column dependencies within the single sample. Notably, traditional tabular transformers only attend across columns and lack a mechanism for row-wise attention to model interactions between different samples—a limitation we directly address in the following section.


\section{Methods}
\label{sec:method}

\begin{figure}[t!]
\centering
\includegraphics[width=\linewidth]{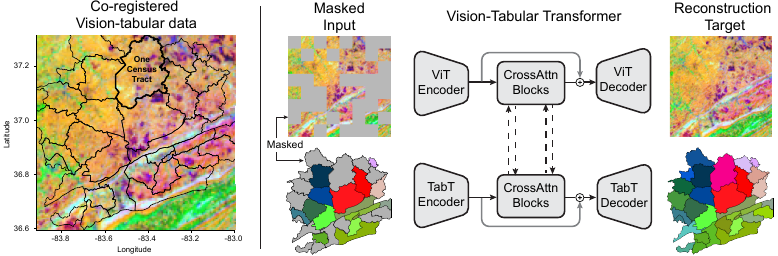}
\caption{\textbf{Paired data and high-level \alg design.} Left: We sample co-registered vision and tabular data. Each tabular row corresponds to a census tract defined by a polygon. Right: We propose \alg, a masked autoencoder framework that jointly trains over paired vision and tabular data. Bilateral cross-attention exchanges spatial and semantic information across modalities. We provide a detailed architectural diagram in \cref{fig:model_architecture}.}
\label{fig:masked_midtrain}
\end{figure}

Consider a pair of spatially co-registered geospatial vision and tabular datasets, where each data point is associated with a longitude $\lambda$ and latitude $\phi$. We extract a local region defined by an $r \times r$ km bounding box centered at location $\mathbf{p}_0=(\lambda_0,\phi_0)$ (\cref{fig:masked_midtrain}). We denote the visual feature image spanning this region $\mathbf{X}_{\mathrm{vis}} \in \mathbb{R}^{H \times W \times C}$, where $H$ and $W$ are the spatial dimensions and $C$ is the number of visual channels. Similarly, we denote the tabular matrix of the $N_{\mathrm{tab}}$ census tracts intersecting this region $\mathbf{X}_{\mathrm{tab}} \in \mathbb{R}^{N_{\mathrm{tab}} \times F}$, where $F$ is the number of tabular features. For each census tract $i$, the corresponding row $\mathbf{x}_{\mathrm{tab},i}$ is associated with a representative location~\footnote{~The representative locations for each census tract is defined by US Census; it is generally at or near the geographic center.} $\mathbf{p}_{\mathrm{tab},i}=(\lambda_{\mathrm{tab},i},\phi_{\mathrm{tab},i})$ and a polygon defining its administrative boundary. A co-registered vision-tabular sample for a given region is therefore the pair $(\mathbf{X}_{\mathrm{vis}},\mathbf{X}_{\mathrm{tab}})$.

We introduce \alg, a vision-tabular transformer designed to learn a rich, joint feature space from this paired data via self-supervised masked autoencoding. We detail the architectural design of \alg in \cref{sec:architecture}, and outline the self-supervised training procedure in \cref{sec:training}.

\subsection{\alg Architecture Design}
\label{sec:architecture}
\begin{figure}[t!]
\centering
\includegraphics[width=0.95\linewidth]{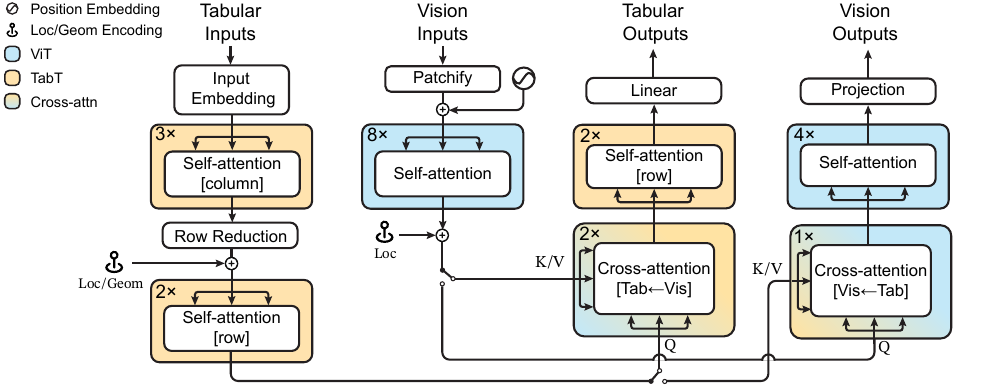}
\caption{\textbf{\alg architecture}. We feed tabular inputs to the row encoder blocks (with column self-attention), followed by row attention blocks (with row self-attention). Vision data goes through standard ViT encoder blocks with positional embeddings. Both vision and tabular tokens receive geospatial positional encodings. This is followed by bilateral cross-attention blocks, which mix vision and tabular tokens. After cross-attention, vision and tabular data are decoded by their respective decoders. We omit residual connections here for clarity.}
\label{fig:model_architecture}
\end{figure}

\alg is a masked vision-tabular autoencoder that jointly reconstructs missing content from visible context (see \cref{fig:masked_midtrain} for an overview, and \cref{fig:model_architecture} for details). The architecture comprises Vision Transformer (ViT) and Tabular Transformer (TabT) encoder-decoders, bridged by bilateral cross-attention blocks prior to decoding. Each modality-specific encoder performs self-attention to facilitate token mixing within its own domain. While we employ a standard ViT for the visual pathway, our TabT incorporates a novel encoder that explicitly supports both column-wise (cross-feature) and row-wise (cross-sample) attention, unlike existing TabTs that restrict attention strictly to features (see \cref{sec:related-work}). We denote the vision token dimension as $D_v$, the tabular column token dimension as $D_{\text{col}}$, and the tabular row token dimension as $D_{t}$.

To ground these inputs, we inject novel geospatial positional encodings based on spatial proximity (for both modalities) and census tract geometry (for tabular data). Bilateral cross-attention blocks subsequently exchange spatial and semantic information across modalities before modality-specific decoders reconstruct the original signals. Complete architectural details are provided in the Supplementary Material. The following subsections detail our geospatial positional encodings, tabular transformer, cross-attention mechanism, and training approach.

\subsubsection{Geospatial Positional Encodings} 
\label{sec:pos_encoding}

\begin{wrapfigure}{r}{0.5\textwidth}
\vspace{-1.0em}
\centering
\includegraphics[width=\linewidth]{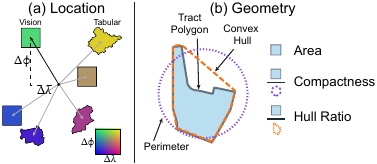}
\caption{\textbf{Location and geometry encodings}. (a) Both vision and tabular tokens are encoded with longitude and latitude offsets from a reference point. (b) Tabular tokens receive additional encoded geometry summaries.}
\label{fig:location_encoding}
\vspace{-0.8em}
\end{wrapfigure}

We devise geospatial positional encodings that capture the spatial proximity of tokens and the approximate polygon geometry of the underlying census tracts. For any coordinate $\mathbf{p}=(\lambda,\phi)$ within a region, we encode its location offset $\Delta \mathbf{p}$ relative to the reference location $\mathbf{p}_0=(\lambda_0, \phi_0)$:
\[
\Delta \mathbf{p} =
\left[
\phi-\phi_0,
(\lambda-\lambda_0)\cos(\phi_0)
\right],
\]
where $\cos(\phi_0)$ scales the east-west displacement by latitude. For each vision patch token $j$, we compute this offset $\Delta \mathbf{p}_{\mathrm{vis},j}$ using its patch-center location $\mathbf{p}_{\mathrm{vis},j}$. For each tabular row token $i$, we compute the offset $\Delta \mathbf{p}_{\mathrm{tab},i}$ using its representative census tract location $\mathbf{p}_{\mathrm{tab},i}$.

For tabular tokens, we also compute a low-dimensional geometry summary based on their census tract polygons (\cref{fig:location_encoding}b). We calculate the area $A_i$, perimeter $P_i$, and convex hull area $H_i$, deriving the log-area $\mu_i=\log(1+A_i)$, compactness $\kappa_i=4\pi A_i/P_i^2$, and convex-hull ratio $\rho_i=A_i/H_i$. Together with the relative location offset, these form the comprehensive tract summary $\mathbf{u}_{\mathrm{tab},i}=[\Delta \mathbf{p}_{\mathrm{tab},i},\mu_i,\kappa_i,\rho_i]$. Finally, learned multilayer perceptrons (MLPs) project the vision location offsets and tabular geometry summaries to match their respective token dimensions:
\[
\mathbf{e}_{\mathrm{vis},j} = f_{\mathrm{vis}}(\Delta \mathbf{p}_{\mathrm{vis},j}) \in \mathbb{R}^{D_v},
\qquad
\mathbf{e}_{\mathrm{tab},i} = f_{\mathrm{tab}}(\mathbf{u}_{\mathrm{tab},i}) \in \mathbb{R}^{D_{t}}.
\]

\subsubsection{Tabular Transformer with Row Attention}
\label{sec:tab_transformer}
Traditional tabular transformers use feature (column) self-attention. We extend this design to incorporate row self-attention, enabling token mixing across census tracts within a local region. Given local tabular data $\mathbf{X}_{\mathrm{tab}} \in \mathbb{R}^{N_{\mathrm{tab}} \times F}$, we first independently apply FT-Transformer encoder blocks (with column attention), producing tokens $\mathbf{Z}_{\mathrm{col}} \in \mathbb{R}^{N_{\mathrm{tab}} \times F \times D_{\text{col}}}$, where $D_{\text{col}}$ is the embedding dimension per column. We concatenate column tokens within each row to obtain vectors in $\mathbb{R}^{N_{\mathrm{tab}} \times F D_{\text{col}}}$ and apply a learned row-reduction projection to dimension $D_t$. We then add the geospatial positional encoding $\mathbf{e}_{\mathrm{tab}}$, and pass the resulting sequence through transformer encoder blocks (with row self-attention), producing the final tabular tokens $\mathbf{Z}_{\mathrm{tab}}\in \mathbb{R}^{N_{\mathrm{tab}} \times D_{t}}$.

\subsubsection{Vision-Tabular Cross-Attention}
\label{sec:cross_attention}

\begin{wrapfigure}{r}{0.5\textwidth}
\vspace{-1.0em}
\centering
\includegraphics[width=\linewidth]{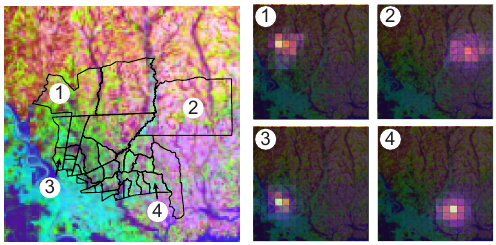}
\caption{\textbf{Spatially localized cross-attention}. For four example tabular tokens, we visualize their cross-attention weights over vision tokens. The learned attention concentrates on nearby visual patches, indicating that cross-modal interactions follow local spatial structure.}
\label{fig:attention}
\vspace{-0.1em}
\end{wrapfigure}

With encoded ViT tokens $\mathbf{Z}_{\mathrm{vis}} \in \mathbb{R}^{N_{\mathrm{vis}} \times D_v}$ and TabT tokens $\mathbf{Z}_{\mathrm{tab}} \in \mathbb{R}^{N_{\mathrm{tab}} \times D_{t}}$, we proceed with bidirectional ViT--TabT cross-attention blocks. Each modality updates its representation using the other as context: vision queries tabular (vis$\leftarrow$tab), and vice versa. Each direction uses separate cross-attention and feed-forward parameters. The two residual updates are applied in parallel within each block, yielding vision-enriched tabular tokens $\mathbf{Z}_{\mathrm{tab}}'$ and tabular-enriched vision tokens $\mathbf{Z}_{\mathrm{vis}}'$. This design preserves modality-specific token spaces while allowing row-level tabular context and patch-level visual context to exchange information. These learned attention weights exhibit strong spatial locality (\cref{fig:attention}), demonstrating that tabular tokens primarily attend to spatially adjacent vision tokens.

\paragraph{Cross-attention bias.} To further encourage localized cross-attention, we inject a spatial bias into the ViT--TabT cross-attention logits. Inspired by ALiBi~\citep{press2022trainshort}, which biases attention by token distance in text, we bias attention by physical spatial distance in kilometers. We compute pairwise distances $d$ between tokens and transform them with a fixed distance function $\phi(d)=\tanh((d_0-d)/\tau)$, where $d_0$ is the zero-crossing distance and $\tau$ is a fixed temperature to control degradation. For each head $h$, we scale this spatial bias by a learnable gain $\alpha_h$. The resulting bias is added directly to the pre-softmax attention scores: $\mathbf{q} \mathbf{k}^\top + \alpha_h\phi(d)$. The vis$\leftarrow$tab and tab$\leftarrow$vis pathways use the same function $\phi(d)$ but learn independent $\alpha_h$ parameters.

\subsection{Self-Supervised Training Objective and Procedure}
\label{sec:training}
We apply the MAE~\citep{he2021masked} training objective to co-registered vision-tabular data $(\mathbf{X}_{\mathrm{vis}},\mathbf{X}_{\mathrm{tab}})$, as seen in \cref{fig:masked_midtrain}. We mask out a fraction of vision patches, as well as a fraction of census tracts. For the tabular data, we randomly mask out entire rows rather than individual features, forcing the model to reconstruct the complete census tract profiles. Reconstruction loss is evaluated only on masked tokens. We use mean-squared error (MSE) for vision and mean absolute error for tabular.

\begin{figure}[h]
\centering
\includegraphics[width=\linewidth]{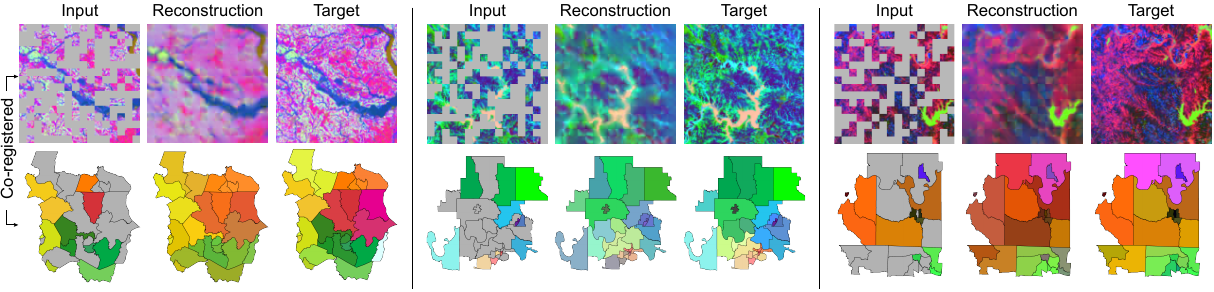}
\caption{\textbf{Masked auto-reconstruction validation results.} Under a mask ratio of $0.5$, we jointly reconstruct missing vision patches and tabular census tracts by using self- and cross-attention to exchange spatial and semantic information across the two modalities.}
\label{fig:mae_result}
\end{figure}

We show example validation results in \cref{fig:mae_result}. For visualization, we used Principal Component Analysis (PCA) to project census tract profiles onto three components, displayed as RGB colors. As in standard vision MAEs, reconstructed image patches tend to be spatially smooth but semantically plausible. Tabular reconstructions show a similar pattern: recovered census profiles appear more spatially smoothed than their targets while preserving broad structural trends, suggesting effective, spatially informed token mixing across modalities.

\section{Experiments} \label{sec:experiments}
\textbf{Datasets.}
We used 64-dimensional annual embedding fields from AlphaEarth \cite{brown2025alphaearth} as our gridded vision data. AlphaEarth provides globally consistent, semantically rich Earth observation (EO) features, making it preferable to low-level spectral inputs. We used all available AlphaEarth embeddings (2017--2024) globally, excluding ocean tiles. We used downsampled overviews (320 m and 640 m resolutions) to better align with the spatial scale of census tracts.

For our tabular dataset, we used the 2018 U.S. Climate Vulnerability Index (CVI) \citep{cviabout}. The CVI offers a data-driven, neighborhood-scale assessment of cumulative environmental, infrastructural, and socioeconomic vulnerability at the census-tract level. This aligns well with our multimodal reasoning objectives. We selected 139 environment-related indicators for training and excluded 44 health-related indicators from the input. We excluded Alaska and Hawaii due to substantial missing features. Due to the vast difference in dataset scales ($73$k census tracts \vs $1.1$M AlphaEarth images), we independently pretrained the ViT backbone.
 
\textbf{ViT Pretraining Implementation.} To prevent data leakage into downstream tasks, we split the gridded data temporally, withholding 2024 for validation and 2018 for testing. We pretrained the ViT using the standard Masked Autoencoder (MAE) framework~\citep{he2021masked}. Our visual backbone is a ViT-L/8 with an encoder dimension of 1024. Because we input analysis-ready AlphaEarth embeddings rather than raw pixels, we reduced the encoder to 8 blocks and the decoder to 4 blocks (dimension 768). Following \citet{he2021masked}, we used a mask ratio of $0.75$ and a batch size of 4096. For ViT pretraining, we augment the standard MSE loss function with an additional cosine distance term, which is appropriate for AlphaEarth embeddings on a 64-dimensional unit sphere.

\textbf{ViT-TabT Joint Training Implementation.}
For joint training, we used co-registered 2018 AlphaEarth (640 m resolution) and CVI datasets with 73k census tracts across the Contiguous United States (CONUS). We withhold Washington state for zero-shot testing and split the remaining data into a 9:1 train-validation set using random 300 km bounding boxes. We train \alg as a joint MAE (\cref{fig:masked_midtrain}), freezing the pre-trained ViT while training the TabT and cross-attention blocks. Each training sample comprises an $80~\text{km}^2$  AlphaEarth crop and the $\sim32$ intersecting census tracts, under a joint mask ratio of $0.5$. TabT has three column self-attention blocks followed by row reduction to a 384-dimensional space and two additional row self-attention blocks. The tab$\leftarrow$vis cross-attention module uses 8 heads and 1 layer, while the vis$\leftarrow$tab module uses 2 heads and 1 layer (see Supplementary).

\subsection{Principal Component Analysis on Learned Geospatial Representation}

\begin{figure}[h]
\centering
\includegraphics[width=\linewidth]{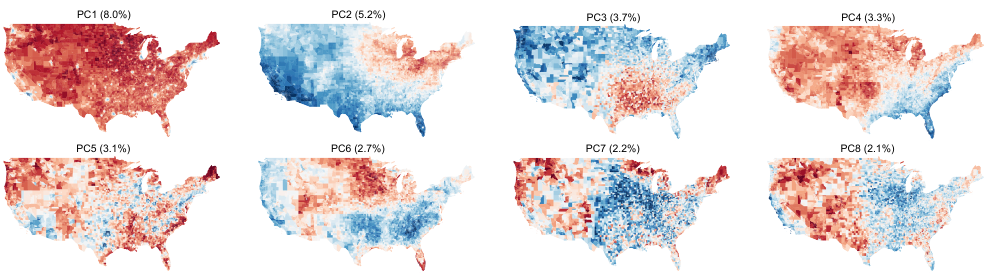}
\caption{\textbf{Principal components of \alg embeddings}. We show the first eight components of vision-enriched tabular embeddings from \alg; percentages indicate explained variance. PC1 shows urban-rural transition, and subsequent PCs show coherent regional clustering.}
\label{fig:pca_embedding}
\end{figure}


We extracted the vision-enriched tabular embeddings $\mathbf{Z}_{\mathrm{tab}}'$ for all census tracts across the CONUS and performed PCA, shown in \cref{fig:pca_embedding}.
The top-1 principal component shows a clear urban-to-rural transition, followed by subsequent components with coherent regional clustering and localized spatial structures. Despite being trained entirely without labels and without global coordinate information, the feature space naturally captures geographically and socioeconomically meaningful variations, suggesting that \alg learns geospatial representations across multiple environmental domains (natural, built, and socioeconomics). In Supplementary, we show that \alg $\mathbf{Z}_{\mathrm{tab}}'$ embedding appears visually cleaner than raw CVI embedding, and appears to be a hybrid between AlphaEarth and CVI, consistent with cross-modal mixing.

\subsection{Linear Probing on Downstream Metrics}

Following representation learning works~\citep{chen2020simple,caron2021emerging,he2021masked}, we evaluate \alg embeddings by their efficacy on downstream predictive tasks. In particular, we extracted \alg's vision-enriched tabular embeddings $\mathbf{Z}_{\mathrm{tab}}'$ for all CONUS census tracts and fit linear regression models (``linear probes'') using scikit-learn~\citep{pedregosa2011scikit} to predict arbitrary downstream variables on a held-out test fraction. 
All evaluation variables correspond to the year 2018.

We evaluated against four baseline feature embeddings: (1) \textbf{CVI}: 139-D tabular inputs with median imputation; (2) \textbf{AlphaEarth}: 64-D vision embeddings averaged per census tract; (3) \textbf{Feature Concat}: Direct concatenation of CVI and AlphaEarth features for each census tract; and (4) \textbf{Late Fusion}: Vision features concatenated with representations from a separately trained tabular MAE (lacking cross-attention), which assesses if the two modalities are complementary under a simple nonlinear fusion (MLP decoder), and with no mixing with adjacent tokens.

\begin{table}[h]
\centering
\footnotesize
\setlength{\tabcolsep}{4pt}
\caption{\textbf{Linear-probe on health outcomes: age-adjusted mortality rate for different underlying causes of death.} We report $R^2$ on randomly held-out test counties. \alg outperforms baselines in all five major underlying causes of death, as well as the composite metric. }
\label{tab:probe_health}
\begin{tabular}{lcccccc}
\toprule
Method & \multicolumn{1}{c}{All} & \multicolumn{1}{c}{Cancer} & \multicolumn{1}{c}{Cardiovascular} & \multicolumn{1}{c}{Diabetes} & \multicolumn{1}{c}{Digestive} & \multicolumn{1}{c}{Respiratory} \\
\midrule
CVI [tab] & 0.726 & 0.529 & 0.394 & 0.273 & 0.521 & 0.500 \\
AlphaEarth [vis] & 0.419 & 0.319 & 0.171 & 0.161 & 0.334 & 0.313 \\
Feature Concat [vis--tab] & 0.770 & 0.616 & 0.478 & 0.315 & 0.598 & 0.573 \\
Late Fusion [vis--tab] & 0.743 & 0.542 & 0.436 & 0.313 & 0.573 & 0.612 \\
\alg (Ours) [tab$\leftarrow$vis] & \textbf{0.801} & \textbf{0.675} & \textbf{0.523} & \textbf{0.548} & \textbf{0.645} & \textbf{0.655} \\
\bottomrule
\end{tabular}
\end{table}

\begin{wraptable}[12]{r}{0.40\textwidth}
\vspace{-1.4em}
\centering
\footnotesize
\setlength{\tabcolsep}{3pt}
\caption{\textbf{Linear probe $R^2$ values on age-adjusted mortality rates on the entire held-out state of Washington.} Results demonstrate that \alg yields superior zero-shot downstream metric performance compared to baselines.}
\label{tab:probe_health_extrapolation}
\begin{tabular}{lc}
\toprule
Method & \multicolumn{1}{c}{All} \\
\midrule
CVI [tab] & 0.569 \\
Feature Concat [vis--tab] & 0.511 \\
Late Fusion [vis--tab] & 0.497 \\
\alg (Ours) [tab$\leftarrow$vis] & \textbf{0.611} \\
\bottomrule
\end{tabular}
\vspace{-0.5em}
\end{wraptable}
\textbf{Prediction of mortality rates.} We fit linear probes to predict age-adjusted mortality rates for various underlying causes of death using CDC WONDER data~\citep{cdcwondermortality2021}. \cref{tab:probe_health} reports the $R^2$ on randomly held-out CONUS counties, testing the models' ability to spatially interpolate. CVI provides a strong baseline, as socioeconomic factors heavily influence health outcomes. Conversely, AlphaEarth performs poorly alone. While Feature Concat and Late Fusion yield marginal improvements over CVI, \alg consistently outperforms all baselines across every mortality category. This suggests that modeling spatially coherent, non-linear relationships between visual and tabular modalities enhances representation quality.

\begin{wraptable}[13]{r}{0.4\textwidth}
\vspace{-1.5em}
\centering
\footnotesize
\setlength{\tabcolsep}{3pt}
\caption{\textbf{Linear probe $R^2$ values on predicting downstream fire hazard measures on held-out counties.} $N_{\mathrm{days}}$ reflects risk frequency, and $I_{\max}$ reflects peak intensity. \alg outperforms baselines on $N_{\mathrm{days}}$ }
\label{tab:probe_fire_risks}
\vspace{0.15em}
\begin{tabular}{lcc}
\toprule
Method & $N_{\mathrm{days}}$ & $I_{\max}$ \\
\midrule
CVI [tab] & 0.671 & 0.866 \\
AlphaEarth [vis] & 0.755 & 0.886 \\
Feature Concat [vis--tab] & 0.788 & 0.886 \\
Late Fusion [vis--tab] & 0.741 & \textbf{0.961} \\
\alg (Ours) [tab$\leftarrow$vis] & \textbf{0.792} & 0.911 \\
\bottomrule
\end{tabular}
\vspace{-0.5em}
\end{wraptable}
To test geographic extrapolation, we evaluated zero-shot transfer to Washington (WA) state, which we entirely held out during joint-training. 
As shown in \cref{tab:probe_health_extrapolation}, zero-shot prediction is substantially harder than interpolation. AlphaEarth fails on mortality extrapolation ($R^2 < 0$, omitted), and simple fusion baselines perform worse than the tabular-only CVI model. In contrast, \alg maintains positive transfer, outperforming all baselines.

\textbf{Prediction of fire hazards.} We further evaluated \alg on the FireCCI51 dataset~\citep{chuvieco2018firecci51} (\cref{tab:probe_fire_risks}) to predict two fire-weather hazards signals: the number of extreme fire risk days ($N_{\mathrm{days}}$) and peak annual intensity ($I_{\max}$). Because fire hazards are primarily driven by the physical environment, AlphaEarth naturally outperforms CVI. However, Feature Concat improves $N_{\mathrm{days}}$ predictions over either unimodal input, indicating that tabular data contributes relevant anthropogenic correlates (e.g., infrastructure, land use). \alg achieves the best performance for extreme fire-weather days, while Late Fusion marginally wins on peak intensity. This suggests that cross-modal attention is highly effective for frequency-based hazards, whereas isolated modality features may better preserve extreme peak values.


\providecommand{\shadecell}[1]{%
  \begingroup
  \setlength{\fboxsep}{1pt}%
  \colorbox{black!10}{\strut #1}%
  \endgroup
}
\providecommand{\ablationpanelcaption}[2]{%
  \vspace{0pt}%
  \raggedright\scriptsize\textbf{#1} #2\par\smallskip
  \centering
}

\begin{table}[h]
\centering
\caption{\textbf{Ablation experiments.} We report linear probe $R^2$ for mortality rates on randomly held-out test counties. We mark default settings in \shadecell{gray}.}
\label{tab:ablation}
\footnotesize
\setlength{\tabcolsep}{3pt}
\renewcommand{\arraystretch}{0.95}

\makebox[\linewidth][c]{%
\begin{tabular}{@{}c@{\hspace{0.015\linewidth}}c@{\hspace{0.015\linewidth}}c@{\hspace{0.015\linewidth}}c@{\hspace{0.015\linewidth}}c@{}}
\begin{minipage}[t]{0.14\linewidth}
\ablationpanelcaption{(a) Tab dim.}{}
\begin{tabular}{@{}cc@{}}
\toprule
dim & $R^2$ \\
\midrule
128 & 0.706 \\
192 & 0.769 \\
256 & 0.748 \\
\shadecell{384} & \shadecell{\textbf{0.801}} \\
512 & 0.787 \\
\bottomrule
\end{tabular}
\end{minipage}
&
\begin{minipage}[t]{0.22\linewidth}
\ablationpanelcaption{(b) Location and geometry.}{}
\begin{tabular}{@{}lc@{}}
\toprule
encoding & $R^2$ \\
\midrule
none & 0.770 \\
geom/loc & 0.788 \\
\shadecell{geom/loc $+$ bias} & \shadecell{\textbf{0.801}} \\
\bottomrule
\end{tabular}
\end{minipage}
&
\begin{minipage}[t]{0.18\linewidth}
\ablationpanelcaption{(c) Tab$\leftarrow$vis cross-attn.}{}
\begin{tabular}{@{}cc@{}}
\toprule
capacity & $R^2$ \\
\midrule
4H1L & 0.771 \\
4H2L & 0.775 \\
\shadecell{8H1L} & \shadecell{\textbf{0.801}} \\
8H2L & 0.777 \\
\bottomrule
\end{tabular}
\end{minipage}
&
\begin{minipage}[t]{0.15\linewidth}
\ablationpanelcaption{(d) Tab mask ratio.}{}
\begin{tabular}{@{}cc@{}}
\toprule
ratio & $R^2$ \\
\midrule
0.25 & 0.791 \\
\shadecell{0.50} & \shadecell{\textbf{0.801}} \\
0.75 & 0.703 \\
\bottomrule
\end{tabular}
\end{minipage}
&
\begin{minipage}[t]{0.16\linewidth}
\ablationpanelcaption{(e) Tab self-attn.}{}
\begin{tabular}{@{}cc@{}}
\toprule
row attn & $R^2$ \\
\midrule
no & 0.767 \\
\shadecell{yes} & \shadecell{\textbf{0.801}} \\
\bottomrule
\end{tabular}
\end{minipage}
\end{tabular}%
}
\vspace{-0.8em}
\end{table}

\paragraph{Ablation studies.}
We ablated \alg's components using the overall CDC WONDER mortality rate on held-out counties (column ``All'' in \cref{tab:probe_health}). As seen in \cref{tab:ablation}(a), a TabT token dimension of 384 best balances information capacity (compressing 139 features and cross-modal vision cues) against overfitting. \cref{tab:ablation}(b) demonstrates that both our geometry/location encodings (\cref{sec:pos_encoding}) and cross-attention spatial biases (\cref{sec:cross_attention}) improve performance; without them, the spatially localized attention patterns observed in \cref{fig:attention} vanish. \cref{tab:ablation}(c) reveals that a wider, shallower cross-attention mechanism (8 heads, 1 layer) outperforms deeper networks, likely preventing the over-mixing of local heterogeneous cues. An intermediate tabular mask ratio of 0.50 proves optimal (\cref{tab:ablation}(d)), balancing task difficulty with sufficient regional context. Finally, \cref{tab:ablation}(e) confirms that row self-attention—enabling interaction across spatially adjacent census tracts—is a critical driver of \alg's predictive success.

\section{Discussion}
\label{sec:discussion}
We presented \alg, a vision-tabular transformer for joint geospatial representation learning from co-registered gridded and tabular datasets. 
Our results underscore that physical and socioeconomic signals contain highly complementary information. 
For instance, \alg's embeddings explained up to $80\%$ of the county-level variance in age-adjusted mortality rates, highlighting the strong predictive signal embedded within a holistic representation of the multimodal environment.
We demonstrated that explicitly modeling interactions between visual environmental signals and structured socioeconomic context yields significantly more transferable representations than unimodal or simple feature-concatenation approaches. Specifically, \alg improved linear-probe performance across downstream health and fire-hazard tasks, while also demonstrating stronger geographic extrapolation to unseen regions. These findings suggest that multimodal geospatial reasoning greatly benefits from architectures that jointly account for spatial proximity and heterogeneous data structures.

\paragraph{Limitations.} First, experiments were restricted to the CONUS using a single tabular dataset (CVI); performance may differ across regions with distinct geographic, demographic, or administrative structures. Second, aggregating tabular variables at the census-tract level may obscure fine-grained within-region heterogeneity and is inherently dependent on the quality of survey-based data collection. Third, while \alg learns geographically meaningful associations, the resulting representations remain correlational and should not be interpreted causally, and such misinterpretation might incur a negative societal impact. Finally, our current framework focuses on static environmental snapshots. Future work could extend \alg to explicitly model temporal dynamics, enabling longitudinal geospatial reasoning over evolving climate, infrastructure, and population conditions globally.

\paragraph{Broader Impact.} This work provides a flexible, self-supervised foundation for broader multimodal geospatial modeling. By learning transferable latent representations without labeled data, it facilitates adaptation to regions with sparse ground-truth annotations. The architecture can naturally be extended to incorporate diverse modalities---such as climate projections, mobility data, electronic health records, or real-time satellite streams---to support large-scale predictive modeling in disaster forecasting, urban planning, climate adaptation, and environmental justice.

\paragraph{Acknowledgments.}
The authors gratefully acknowledge support for this research from the National Science Foundation (NSF) under award IIS-2107313. Any opinions, findings, conclusions, or recommendations expressed in this paper are those of the authors and do not necessarily reflect the views of the sponsors.


{\small
\bibliographystyle{IEEEtranN}
\bibliography{references_zotero_yuhao,references}
}

\clearpage
\appendix
\begin{center}
{\Large\bfseries Supplementary Material}\\[0.5em]
{\Large\bfseries \papertitle}
\end{center}
\vspace{1em}
\crefalias{section}{appendix}
\section{Geospatial Dataset Details}
\label{sec:appendix-data}

\subsection{AlphaEarth Vision Data}
We use AlphaEarth Foundations~\citep{brown2025alphaearth} as the gridded vision modality.
AlphaEarth provides annual, analysis-ready embedding fields in which each grid cell is represented by a
64-dimensional vector summarizing Earth-observation context.
Unlike raw optical, radar, or climate bands, these channels are latent embedding dimensions and do not have
individual physical units; visualizations of AlphaEarth fields therefore use PCA-RGB projections of the
64-dimensional vectors.
We use all available annual fields from 2017--2024 for ViT pretraining, excluding ocean and no-data tiles.
To match the spatial scale of census-tract-level CVI attributes, we use downsampled AlphaEarth overviews at
320 m and 640 m resolution.
Joint training uses the 2018 CONUS AlphaEarth field co-registered with census-tract CVI rows.
For the vision-only linear-probe baseline, we average the 64-dimensional AlphaEarth vectors within each
census tract before fitting the downstream probe.

\subsection{CVI Tabular Environmental Data}
We use the U.S. Climate Vulnerability Index (CVI)~\citep{cviabout} as the tabular environmental modality.
CVI provides census-tract-level indicators covering socioeconomic vulnerability, infrastructure, healthcare
access, environmental exposures, pollution sources, and extreme-event risks.
We use the variables marked as \texttt{Input} in the CVI definition workbook as tabular covariates for
self-supervised training, and withhold the variables marked as \texttt{Outcome} so that health indicators are
not used as reconstruction targets.
We exclude Alaska and Hawaii because many CVI columns are missing for those states, and use the remaining
contiguous United States census tracts for joint training and evaluation.
Table~\ref{tab:cvi-input-definitions} lists the CVI input variables used as tabular environmental data.

\footnotesize
\setlength{\tabcolsep}{4pt}
\renewcommand{\multirowsetup}{\raggedright}

\begin{longtable}{>{\raggedright\arraybackslash}m{0.2\linewidth}>{\itshape}p{0.10\linewidth}p{0.58\linewidth}}
\caption{Tabular exposome inputs in the CVI definition workbook.\label{tab:cvi-input-definitions}}\\
\toprule
\textbf{Subdomain} & \textbf{\upshape Key} & \textbf{Indicator} \\
\midrule
\endfirsthead
\toprule
\textbf{Subdomain} & \textbf{\upshape Key} & \textbf{Indicator} \\
\midrule
\endhead
\midrule
\multicolumn{3}{r}{Continued on next page} \\
\midrule
\endfoot
\bottomrule
\endlastfoot
\multicolumn{3}{l}{\textbf{\textit{Domain: Socio-economic}}} \\
\midrule
\multirow[c]{8}{=}{Socioeconomic Stressors} & BlwPv & Below Poverty \\
 & Unmpl & Unemployed \\
 & LwInc & Low Income \\
 & NHgSD & No High School Diploma \\
 & HmcdR & Homicide Rate \\
 & GnVln & Gun Violence \\
 & RlgsO & Religious Organizations \\
 & CvcSO & Civic and Social Organizations \\
\cmidrule(lr){1-3}
\multirow[c]{5}{=}{Housing Composition \& Disability} & Ag65O & Aged 65 or Older \\
 & Ag17Y & Aged 17 or Younger \\
 & CvlnD & Civilian with a Disability \\
 & SngPH & Single-Parent Households \\
 & FstrC & Foster Children \\
\cmidrule(lr){1-3}
\multirow[c]{8}{=}{Minority Status \& Language} & Mnrty & Minority \\
 & SpELW & Speaks English Less than Well \\
 & UndcP & Undocumented Population \\
 & HtCrm & Hate Crimes \\
 & PrsnP & Prison Population \\
 & Rdlnn & Redlining \\
 & HmlsP & Homeless Population \\
 & VtrnP & Veterans Population \\
\cmidrule(lr){1-3}
\multirow[c]{7}{=}{Housing \& Transportation} & MltUS & Multi-Unit Structures \\
 & MblHm & Mobile Homes \\
 & Crwdn & Crowding \\
 & NVhcl & No Vehicle \\
 & GrpQr & Group Quarters \\
 & HsnFR & Housing Foreclosure Risk \\
 & PHUBB & Percent of Housing Units Built Between 1940-1969 as of 2015-2019 \\
\cmidrule(lr){1-3}
\multirow[c]{5}{=}{Costs of Climate Disasters} & FEMAH & FEMA Hazard Mitigation Grants \\
 & Fldnrtp & Flooding risk to properties \\
 & Wldfr & Wildfire risk to properties \\
 & Pttb2 & Property taxes expected to be lost by 2045 due to chronic inundation \\
 & Cstfc & Cost of climate disasters \\
\cmidrule(lr){1-3}
\multirow[c]{6}{=}{Productivity Losses} & HRJPC & High-Risk Jobs Productivity (\% Change) \\
 & Yldsc & Yields (\% change) \\
 & Owwdr & Outdoor workers - work days at risk per year \\
 & EALAV & Expected Annual Loss - Agriculture Value \\
 & EALBV & Expected Annual Loss - Building Value \\
 & EALPE & Expected Annual Loss - Population Equivalence \\
\cmidrule(lr){1-3}
\multirow[c]{4}{=}{Transition Risks} & RsdEE & Residential Energy Expenditures (\% change) \\
 & ShrJA & Share of Jobs in Agriculture \\
 & Scdby & State energy-related carbon dioxide emissions by year \\
 & MthnE & Methane Emissions \\
\cmidrule(lr){1-3}
\multirow[c]{2}{=}{Social Stressors} & PrprC & Property Crimes (\% change) \\
 & VlntC & Violent Crimes (\% change) \\
\midrule
\multicolumn{3}{l}{\textbf{\textit{Domain: Infrastructure}}} \\
\midrule
\multirow[c]{8}{=}{Transportation} & Dlycp & Delay (congestion) per capita/census tract \\
 & Fldnrtr & Flooding risk to roads \\
 & Lnmls & Lane miles per capita \\
 & RdQlM & Road Quality and Maintenance \\
 & PblTP & Public Transit Performance \\
 & BrdQM & Bridge Quality and Maintenance \\
 & Wlkbl & Walkability \\
 & Bkblt & Bikability \\
\cmidrule(lr){1-3}
\multirow[c]{3}{=}{Energy} & RsECB & Residential Energy Cost Burden \\
 & Shfff & Share of energy from fossil fuels \\
 & EVChS & EV Charging Stations \\
\cmidrule(lr){1-3}
\multirow[c]{4}{=}{Food, Water, and Waste Management} & MRFEI & Modified Retail Food Environment Index \\
 & FdIns & Food Insecurity \\
 & AccHF & Access to Healthy Foods \\
 & IndrP & Indoor Plumbing \\
\cmidrule(lr){1-3}
\multirow[c]{2}{=}{Communications} & PoHwn & Percent of Household with no internet access \\
 & Phwsb & Percent of household with smartphone but no other device. \\
\cmidrule(lr){1-3}
\multirow[c]{4}{=}{Financial Services} & PrcUH & Percent of Unbanked Households \\
 & Pydyl & Payday lending rank \\
 & HsngAfr & Housing Affordability (renters) \\
 & HsngAfo & Housing Affordability (owners) \\
\cmidrule(lr){1-3}
\multirow[c]{5}{=}{Governance} & TBMRE & Tax Base: Median Real Estate Taxes Paid \\
 & VT202 & Voter Turnout 2020 \\
 & PblLL & Public Library Locations \\
 & HUDPH & HUD Public Housing \\
 & AffHU & Aggregate funding amount for HUD grants \\
\midrule
\multicolumn{3}{l}{\textbf{\textit{Domain: Healthcare}}} \\
\midrule
\multirow[c]{5}{=}{Access to Care} & PrxNH & Proximity to Nursing Homes \\
 & NHBp1 & Number of Hospital Beds per 10,000 people \\
 & MdcUA & Medically Underserved Areas \\
 & CrLHI & Current Lack of Health Insurance \\
 & Prxmt & Proximity to hospitals \\
\midrule
\multicolumn{3}{l}{\textbf{\textit{Domain: Environment}}} \\
\midrule
\multirow[c]{8}{=}{Transportation Sources} & Ttvmt & Total vehicle miles traveled per capita \\
 & Psvmt & Passenger vehicle miles traveled per capita \\
 & Trvmt & Truck vehicle miles traveled per capita \\
 & HDVvm & Heavy Duty Vehicle vehicle miles traveled per capita \\
 & PrxmP & Proximity to Ports \\
 & RlCrs & Rail Crossings \\
 & TrfPV & Traffic Proximity and Volume \\
 & NtTNM & National Transportation Noise Map \\
\cmidrule(lr){1-3}
\multirow[c]{14}{=}{Exposures \& Risks} & RSEIR & Risk-Screening Environmental Indicators (RSEI) \\
 & ArTxRs & Air Tox Respiratory \\
 & ArTxN & Air Tox Neurological \\
 & ArTxL & Air Tox Liver \\
 & ArTxD & Air Tox Developmental \\
 & ArTxRp & Air Tox Reproductive \\
 & ArTxK & Air Tox Kidney \\
 & ArTxI & Air Tox Immunological \\
 & ArTxT & Air Tox Thyroid \\
 & ATTCR & Air Tox Total Cancer Risk \\
 & BlckC & Black Carbon \\
 & Agrcl & Agricultural pesticides \\
 & LPhb1 & Lead Paint: \% housing units built before 1960 \\
 & Ldndw & Lead in drinking water violations \\
\cmidrule(lr){1-3}
\multirow[c]{14}{=}{Pollution Sources} & SprfS & Superfund Sites \\
 & Brwnf & Brownfields \\
 & STRSE & Stream Toxicity Risk-Screening Environmental Indicators (RSEI) \\
 & Ptfpm & Proximity to facilities participating in air markets \\
 & NPLst & NPL sites \\
 & HWMFT & Hazardous Waste Management Facilities (TSDFs) \\
 & HWGnI & Hazardous Waste Generator/Incinerators \\
 & FclEV & Facilities with Enforcement or Violation \\
 & Lndfl & Landfills \\
 & TSCAF & TSCA Facilities \\
 & RsMPF & Risk Management Plan Facilities \\
 & ChmcM & Chemical Manufacturers \\
 & MtlRc & Metal Recyclers \\
 & AcOGW & Active Oil and Gas Wells \\
\cmidrule(lr){1-3}
\multirow[c]{3}{=}{Criteria Air Pollutants} & APM25 & Annual average PM2.5 concentrations \\
 & NO2cn & NO2 concentration \\
 & Ozncn & Ozone concentration \\
\cmidrule(lr){1-3}
\multirow[c]{4}{=}{Land Use} & PrksG & Parks and Greenspace \\
 & ImprS & Impermeable Surfaces \\
 & FrsLC & Forest Land Cover \\
 & NtvAL & Native American Lands \\
\midrule
\multicolumn{3}{l}{\textbf{\textit{Domain: Extreme Events}}} \\
\midrule
\multirow[c]{7}{=}{Temperature} & ClWAF & Cold Wave - Annualized Frequency \\
 & Dwm3C & Days with maximum temperature above 35C \\
 & Dwm40 & Days with maximum temperature above 40C \\
 & FrstD & Frost Days \\
 & Mxmmm & Maximum of maximum temperatures \\
 & Mntmp & Mean temperature \\
 & UHIEH & Urban Heat Island Extreme Heat Days \\
\cmidrule(lr){1-3}
\multirow[c]{2}{=}{Droughts} & DrgAF & Drought - Annualized Frequency \\
 & CnsDD & Consecutive Dry Days \\
\cmidrule(lr){1-3}
\multirow[c]{2}{=}{Wildfires} & WldAF & Wildfire - Annualized Frequency \\
 & SPM25 & Surface PM2.5 \\
\cmidrule(lr){1-3}
\multirow[c]{3}{=}{Precipitation} & Snwfl & Snowfall \\
 & StnPI & Standardized Precip Index \\
 & TtlPr & Total Precipitation \\
\cmidrule(lr){1-3}
\multirow[c]{3}{=}{Flooding} & CsFAF & Coastal Flooding - Annualized Frequency \\
 & RvFAF & Riverine Flooding - Annualized Frequency \\
 & SLvlR & Sea Level Rise \\
\cmidrule(lr){1-3}
\multirow[c]{3}{=}{Storms} & HrrAF & Hurricane - Annualized Frequency \\
 & TrnAF & Tornado - Annualized Frequency \\
 & WnWAF & Winter Weather - Annualized Frequency \\
\end{longtable}
\normalsize

\paragraph{Dropped CVI input variable.}
We additionally drop the environmental input variable C19VR (Covid-19 vaccination rates) because we observed
unreasonable outlier readings in some Texas counties, including percentile reading above 1000\%.

\paragraph{Excluded health indicators.}
We exclude the following CVI outcome variables from self-supervised training:
\textit{overall physical health}: LfExp (life expectancy), SlRPH (self-reported physical health);
\textit{mental health and deaths of despair}: SlRMH (self-reported mental health), DOD10 (drug overdose
deaths), AlchA (alcohol abuse), ScdRt (suicide rates);
\textit{chronic disease}: CrrnD (current diabetes), CrrAA (current adult asthma), Strok (stroke), COPD,
CHD, Cancr (cancer);
\textit{chronic disease prevention}: HghBP (high blood pressure), ChlsS (cholesterol screening), RtnDV
(routine doctor visit), Clnsc (colonoscopy), Mmmgr (mammogram), OlMPS (older men preventive screening),
OlWPS (older women preventive screening), DntlE (dental exams);
\textit{infectious diseases}: COVID (COVID-19 deaths), HpttA (hepatitis A), HpttB (hepatitis B), HIV,
Chlmy (chlamydia), Gnrrh (gonorrhea), Syphl (syphilis), Adsld (Aedes albopictus dengue transmission
increase), Adsgd (Aedes aegypti dengue transmission increase), Adszt (Aedes aegypti zika transmission
increase);
\textit{child and maternal health}: InfnM (infant mortality), Lwbrt (low birthweight), Prtrm (pre-term
birth), ChldA (childhood asthma), Tnbrt (teen births), ADHDP (ADHD prevalence), ADHDT (ADHD treatment),
ChldM (child mortality), FRPSL (free or reduced price school lunch);
\textit{climate- and pollution-related health outcomes}: Tmprt (temperature-related mortality), Dthsf
(deaths from climate disasters), IPMC6 (increased PM$_{2.5}$ mortality--CVD, ages 65+), IncOm (increased
ozone mortality), and Incic (increase in childhood asthma incidence).

\section{Training}
\label{sec:appendix-training}

\begin{figure}[h]
\centering
\includegraphics[width=\linewidth]{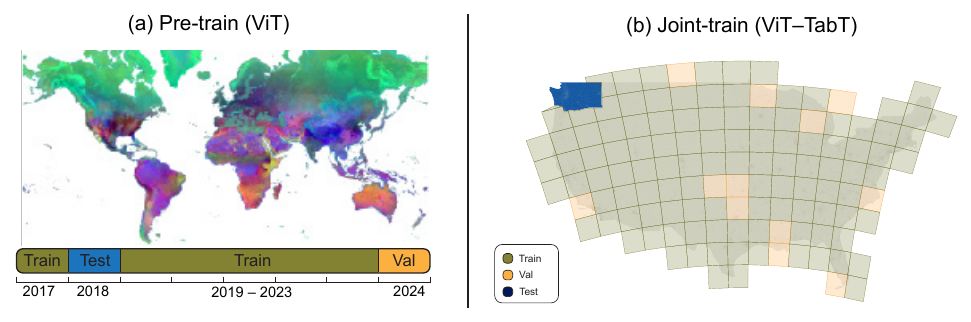}
\caption{\textbf{Training data splits}.}
\label{fig:data_split}
\end{figure}

\paragraph{Additional details on data splits}
\cref{fig:data_split} provides further details on data splits. 
\cref{fig:data_split}a shows that Pretraining data for ViT contains 8 years of global AlphaEarth embedding. We withhold 2018 for test, and 2024 for validation. The map shown here is AlphaEarth embedding, whose color is PCA-RGB embedding, not representative of the data split. \cref{fig:data_split}b shows Joint-training data for \alg comprises 2018 AlphaEarth and CVI tabular datapoints across the CONUS. Here we withhold WA for test, and split train/val via random $300~\text{km}^2$ bounding boxes.

\paragraph{Additional details on model.}
Table~\ref{tab:training-summary} summarizes the model components trained at each stage, the attention
mechanism used by each component, the number of trainable parameters, and the hardware used for training.
The ViT pretraining stage was run on an institutional SLURM cluster using 4$\times$ NVIDIA B200 GPUs; it completed in 35 hours, with GPU memory usage of approximately 80 GB per GPU, on a node with Intel Xeon Platinum 8570 CPUs and 2 TB DRAM. The joint-training stage was run on an internal research-lab cluster using 1$\times$ NVIDIA A100 GPU; it completed in 8 hours, with approximately 70 GB GPU memory usage, on a node with an Intel Xeon Platinum 8362 CPU @ 2.80 GHz and 1 TB DRAM. The full research project did not require additional computing beyond the experiments reported in this paper.

\paragraph{ViT pretraining hyperparameters.}
We pretrain the ViT with AdamW ($\beta_1=0.9$, $\beta_2=0.95$), weight decay $0.05$, base learning rate $1.5\times10^{-4}$, and a warmup-cosine schedule with 40 warmup epochs and a 1000-epoch horizon. Training uses per-GPU batch size 1024 on 4 GPUs (effective batch size 4096), mixed precision, gradient accumulation 1, mask ratio $0.75$, and random seed/split seed 42. The pretraining objective is $L_{\mathrm{pre}}=L_{\mathrm{MSE}}+\beta L_{\mathrm{cos}}$ with $\beta=1$.

\paragraph{\alg joint-training hyperparameters.}
We train \alg with AdamW ($\beta_1=0.9$, $\beta_2=0.95$), learning rate $3\times10^{-4}$, weight decay $0.04$, batch size 256, and a warmup-cosine schedule with 2 warmup epochs and a 100-epoch horizon. Each epoch samples 100,000 local regions; we use random seed 42, random-square train/validation split seed 1339, and vision and tabular mask ratios of $0.5$. The joint MAE objective uses a 1:1 reconstruction weighting, $L_{\mathrm{joint}}=L_{\mathrm{vis}}+\lambda L_{\mathrm{tab}}$ with $\lambda=1$. For the spatial attention bias, we set $d_0=10$ km and $\tau=25$ km, with learnable gains initialized to $1.0$, using distances between vision patch centers and census-tract representative points.

\begin{table*}[h]
\caption{\textbf{Training-stage summary.}}
\label{tab:training-summary}
\small
\centering
\setlength{\tabcolsep}{4pt}
\begin{tabular}{>{\raggedright\arraybackslash}m{0.13\textwidth} >{\raggedright\arraybackslash}m{0.30\textwidth} >{\raggedright\arraybackslash}m{0.23\textwidth} >{\raggedright\arraybackslash}m{0.10\textwidth} >{\raggedright\arraybackslash}m{0.10\textwidth}}
\toprule
\textbf{Stage} & \textbf{Model} & \textbf{Attention Type} & \textbf{$|\theta_{\mathrm{train}}|$} & \textbf{Hardware} \\
\midrule
Pretraining & Vision Transformer (ViT) & Self-attention & 137 M & 4$\times$ B200 \\
\midrule
\multirow[c]{3}{*}{Joint-training} & Tabular Row Encoder Blocks & Self-attention [column] & 416 K & 1$\times$ A100 \\
 & Tabular Transformer Blocks\textsuperscript{a} & Self-attention [row] & 11.6 M & 1$\times$ A100 \\
 & Cross Attention Blocks & Cross-attention & 27.1 M & 1$\times$ A100 \\
\bottomrule
\end{tabular}
\vspace{0.25em}
\parbox{0.86\textwidth}{\footnotesize\textsuperscript{a} Excluding row encoder blocks.}
\normalsize
\end{table*}

\begin{figure}[h]
\centering
\includegraphics[width=\linewidth]{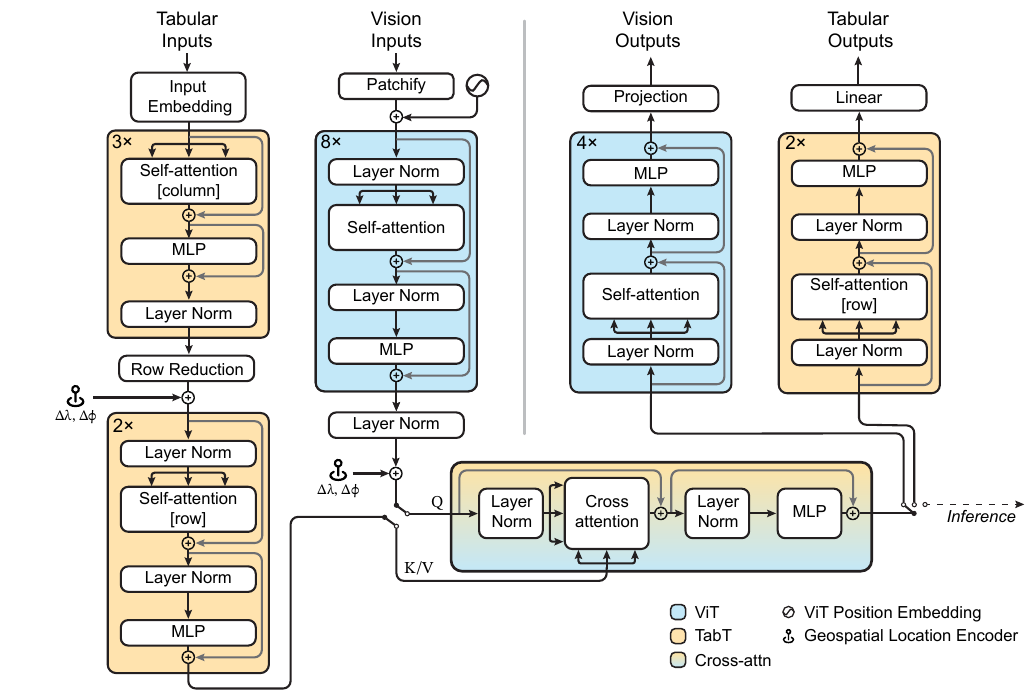}
\caption{\textbf{Detailed \alg architecture}.}
\label{fig:model_architecture_detailed}
\end{figure}

\cref{fig:model_architecture_detailed} offers a more detailed view of \alg than the simplified schematic in \cref{fig:model_architecture}, showing additional details, like residual connections and MLP blocks. For conciseness, only one cross-attention block is drawn; vit$\leftarrow$tabt and tabt$\leftarrow$vit have their own attention blocks and do not share attention weights

\FloatBarrier
\section{Evaluation}

\begin{figure}[h]
\centering
\includegraphics[width=\linewidth]{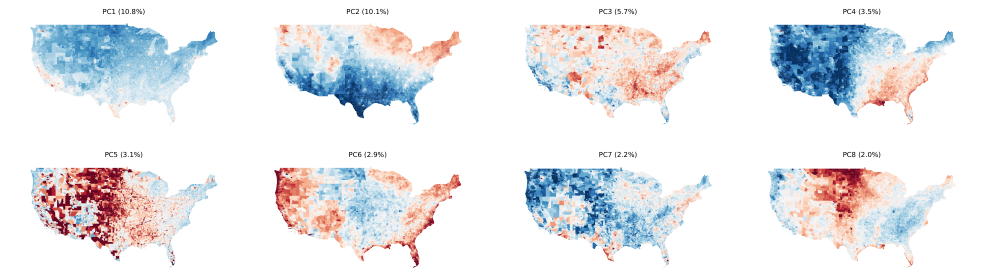}
\caption{\textbf{Principal components of CVI tabular inputs}.}
\label{fig:embedding_pca_cvi}
\end{figure}
\begin{figure}[h]
\centering
\includegraphics[width=\linewidth]{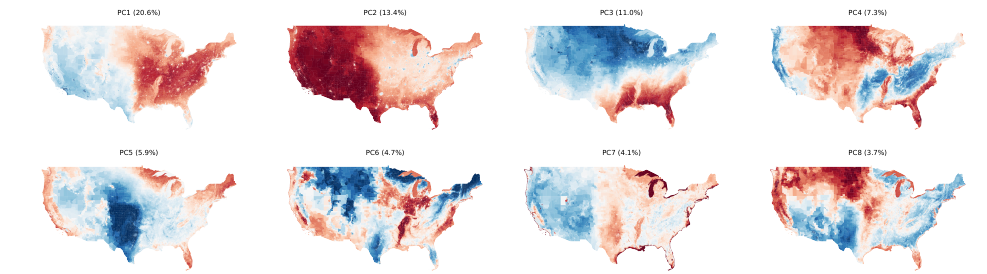}
\caption{\textbf{Principal components of AlphaEarth embeddings}.}
\label{fig:embedding_pca_afe}
\end{figure}

\paragraph{Additional details on PCA.}
Following \cref{fig:pca_embedding}, we also show PCA maps of input CVI tabular features in \cref{fig:embedding_pca_cvi}, and AlphaEarth embeddings (aggregated to census tracts) in \cref{fig:embedding_pca_afe}.
CVI embedding appears to be noisier than \alg's, with outlier counties in both PC1 (random rural splot in WA having the same trend as dense urban centers) and PC2.
AlphaEarth's embedding seems to have a larger emphasis on the natural environment, where PC1 is dominated by east-west separation, rather than urban-rural transition.

\crefalias{section}{section}


\end{document}